\pdfoutput=1
\documentclass{article}

\usepackage[preprint]{corl_2026} %
\hypersetup{pdftitle={STRIDE: Learning Humanoid Locomotion in Minutes with Acceleration-Based Rewards}, pdfauthor={Nico Bohlinger, Jan Peters}}

\usepackage{graphicx}
\usepackage{booktabs}
\usepackage{amsmath}
\usepackage{amssymb}
\usepackage{dsfont}

\newcommand{\minFullGOne}{13}

\newcommand{\minNoSymGOne}{21}

\newcommand{\minNoSymTOne}{18}

\newcommand{\minNoGainGOne}{22}

\newcommand{\minNoGainTOne}{66}

\newcommand{\minDefaultGOne}{49}

\newcommand{\minDefaultTOne}{82}

\newcommand{\minUTDGOne}{30}

\newcommand{\minUTDTOne}{53}

\newcommand{\fallsFull}{440}

\newcommand{\maxMinFull}{14}
\newcommand{\minMinFull}{12}

\newcommand{\minDefault}{56}

\newcommand{\fallsRedFullVsDefault}{81}
\newcommand{\finalVelFull}{0.48}

\newcommand{\fallsBeforeWalkMedianPct}{99}

\newcommand{\budgetMin}{83}
\newcommand{\budgetK}{250k}
\newcommand{\controlHz}{50}

\newcommand{\method}{STRIDE}
\DeclareMathOperator{\clip}{clip}

\title{STRIDE: Learning Humanoid Locomotion\\in Minutes with Acceleration-Based Rewards}

\author{
  Nico Bohlinger$^{1}$ \qquad Jan Peters$^{1,2,3,4}$\\
  $^{1}$Technical University of Darmstadt, Germany \qquad $^{2}$Robotics Institute Germany (RIG)\\
  $^{3}$German Research Center for AI (DFKI) \qquad $^{4}$hessian.AI
}

\begin{document}
\maketitle

\begin{figure}[h]
  \centering
  \includegraphics[width=\linewidth, trim=0 90 0 30, clip]{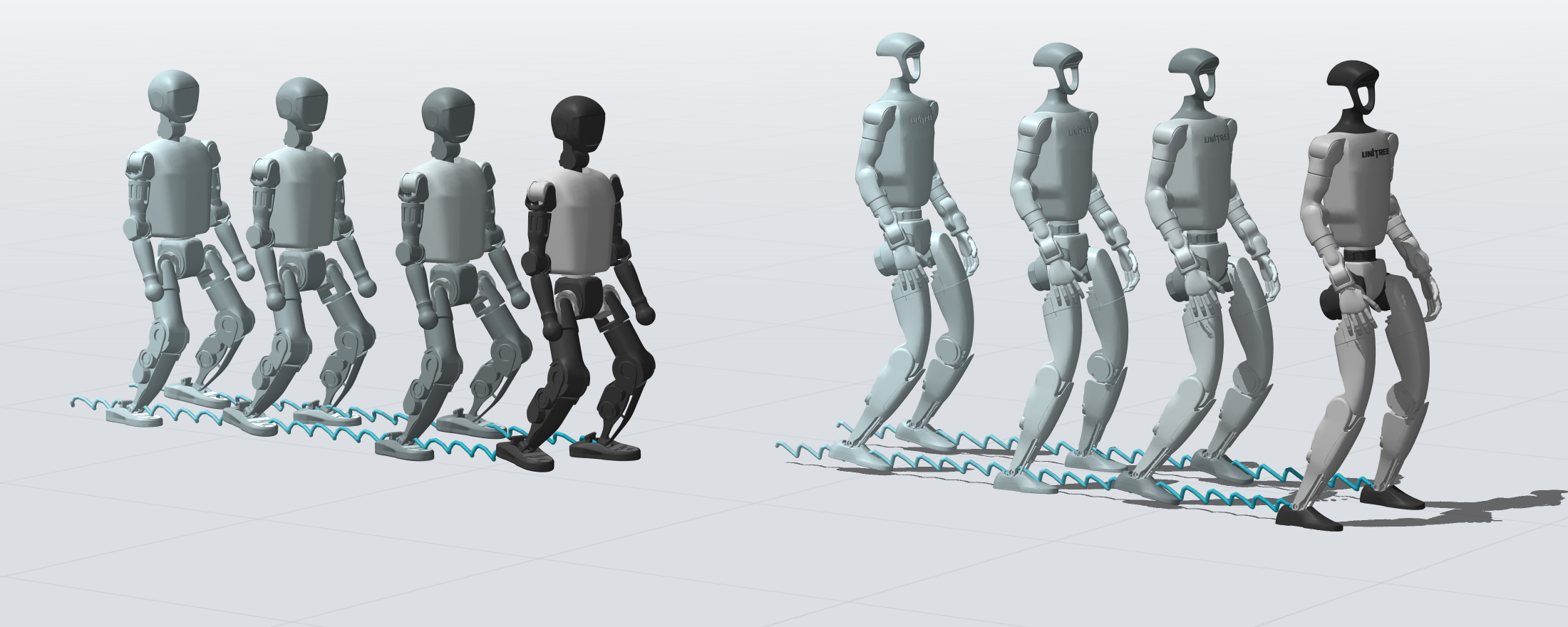}
  \caption{The Booster T1 (left) and the Unitree G1 (right) walk with policies that \method{} learned in \maxMinFull{} minutes worth of collected samples.}
  \label{fig:teaser}
\end{figure}

\begin{abstract}
Learning locomotion directly on a humanoid robot is a promising direction to avoid the sim-to-real gap.
However, locomotion rewards in Reinforcement Learning (RL) typically track the base velocity, which cannot be measured onboard without an external motion capture system or well-calibrated state estimator.
This is why we introduce \method{}, which learns humanoid locomotion from minutes of samples with a reward that only requires the onboard Inertial Measurement Unit (IMU).
Instead of tracking the base velocity, \method{} rewards the robot for following a commanded acceleration profile.
To learn efficiently from this reward, we use the off-policy RL algorithm XQC and combine a high gain setting for the joint position controllers with left-right symmetry augmentation.
In simulation, \method{} learns to walk on the Unitree G1 and Booster T1 humanoids within \maxMinFull{} minutes of samples.
Compared to learning without these two components, it reduces the falls during training by \fallsRedFullVsDefault\%, while without them the robot learns only in over an hour.
\end{abstract}


\section{Introduction}
\label{sec:intro}

Reinforcement Learning (RL) has become the standard approach for legged locomotion~\citep{hwangbo2019agile, radosavovic2024humanoid, stasica2025bridge}, where policies are trained in thousands of parallel environments in simulation and then transferred to the real robot~\citep{rudin2022minutes, bohlinger2025epvf, bohlinger2024onepolicy, bohlinger2025multiembodiment}.
This transfer requires accurate models, extensive domain randomization and often an adaptation to the physical robot~\citep{bohlinger2026active, li2026rapid}.
Learning directly on the robot avoids this gap, and quadrupeds already learn to walk within minutes of real-world interaction~\citep{smith2023walk, bohlinger2025gaitineight}.
Yet, two properties of humanoids make learning on the robot especially difficult.
First, the locomotion rewards typically used in prior work track the base velocity, which has to come from a motion capture system or from a well-calibrated state estimator~\citep{aditya2025robust}.
But even such a state-estimator can still introduce errors and delays, or systematically underestimate the velocity~\citep{bohlinger2025gaitineight}.
Second, bipeds are inherently unstable, so they need large amounts of samples to learn from and fall frequently.
Every fall on a real robot requires a reset and wears down the hardware.

This is why we propose \method{}, which learns humanoid locomotion from minutes of samples with a reward that only requires the onboard Inertial Measurement Unit (IMU).
Instead of tracking the base velocity, \method{} rewards the robot for following a commanded acceleration profile and for holding the velocity that it estimates by integrating its own acceleration.
To learn efficiently from this reward, we use a high gain setting for the joint position controllers and train an off-policy RL algorithm on symmetry-augmented copies of the collected transitions.
In simulation, \method{} learns to walk on the Unitree G1 and Booster T1 humanoids within \maxMinFull{} minutes worth of samples.
Compared to learning with a low gain setting and without symmetries, it reduces the falls during training by \fallsRedFullVsDefault\%, while only half of the runs without these two components learn to walk within \budgetMin{} minutes at all.

\section{Related Work}
\label{sec:related}

Locomotion has been learned directly on quadrupeds with maximum entropy RL~\citep{haarnoja2019walk} and with world models~\citep{wu2022daydreamer, li2025robotic}, and more sample-efficient off-policy algorithms reduced the training time to minutes~\citep{smith2023walk, bohlinger2025gaitineight}.
This efficiency builds on better-conditioned critic learning~\citep{bhatt2024crossq, palenicek2025xqc} and higher Update-To-Data (UTD) ratios~\citep{doro2023replayratio}, while the action space and the low-level controller also influence how fast a policy learns~\citep{smith2024grow}.
Structural knowledge reduces the number of required samples further, such as the left-right symmetry of the robot~\citep{mittal2024symmetry} or the periodic gait prior of a Central Pattern Generator (CPG)~\citep{bellegarda2022cpgrl}.
\method{} combines these ideas for humanoids, which are considerably less stable than quadrupeds, and removes the need for a base velocity estimate.
Finally, embodiment-aware architectures scale a single policy to many different robots in simulation~\citep{bohlinger2024onepolicy, bohlinger2025multiembodiment, ai2025embodiment, bohlinger2026shape}, and we take a first step toward learning such a shared policy from the limited data of a few robots.

\section{Learning Humanoid Locomotion Efficiently}
\label{sec:method}

We propose to learn humanoid locomotion with a reward that is computed only from the IMU and to make learning from this reward efficient enough for the real world.
We formulate the task as a Markov Decision Process, in which the policy $\pi_\theta(a_t \mid o_t)$ maps the observation $o_t$ to position targets $a_t$ for the leg joints at every control step $t$.
The observation consists of the joint positions and velocities, the previous action, and the linear acceleration, the angular velocity and the gravity direction from the IMU.
It also contains the commanded velocity, the reference velocity and the goal acceleration, which we introduce in Section~\ref{sec:reward} and which let the policy see what it is rewarded for.
Neither the observation nor the reward requires a measured base velocity.

\subsection{Acceleration-Based Reward}
\label{sec:reward}

An IMU does not measure the velocity of the robot but the linear acceleration $\alpha_t$ of its base, so we formulate the locomotion task in terms of accelerations in the horizontal plane of the IMU frame.
To turn a commanded velocity $v^\mathrm{c}$ into an acceleration profile that the robot can follow, we define a reference velocity $v^\mathrm{r}_t$ that approaches the commanded velocity with a bounded acceleration.
At the same time, the robot keeps an onboard velocity estimate $\hat{v}_t$ by integrating its acceleration as
\begin{equation}
  v^\mathrm{r}_t = v^\mathrm{r}_{t-1} + \clip_{\alpha_\mathrm{max}\Delta t}\!\left(v^\mathrm{c} - v^\mathrm{r}_{t-1}\right) \qquad \text{and} \qquad
  \hat{v}_t = \hat{v}_{t-1} + \alpha_t\,\Delta t,
  \label{eq:reference}
\end{equation}
where $\Delta t$ is the control period and $\alpha_\mathrm{max}$ is the maximum acceleration.
A Proportional-Integral (PI) law then computes a goal acceleration $g_t$ that drives the estimate toward the reference as
\begin{equation}
  g_t = \clip_{\alpha_\mathrm{max}}\!\Big(k_\mathrm{P}\, e_t + k_\mathrm{I} \sum\nolimits_{i \le t} e_i\,\Delta t\Big) \qquad \text{with} \qquad e_t = v^\mathrm{r}_t - \hat{v}_t,
  \label{eq:goal}
\end{equation}
where $k_\mathrm{P}$ and $k_\mathrm{I}$ are the gains.
The integral term keeps increasing the goal acceleration as long as there is a velocity error, so the robot does not stay below the commanded velocity.
Walking accelerates and decelerates the base within every step the robot takes, and a reward only on the instantaneous acceleration would penalize this natural variation.
Therefore, we low-pass filter the measured and the goal acceleration with the time constant $\tau$ and note them as $\bar{\alpha}_t$ and $\bar{g}_t$.
We then define the acceleration reward and the velocity hold reward as
\begin{equation}
  r^\mathrm{acc}_t = \frac{\bar{\alpha}_t\,(2\bar{g}_t - \bar{\alpha}_t)}{\alpha_\mathrm{max}^2} \qquad \text{and} \qquad
  r^\mathrm{hold}_t = \exp\!\left(-\frac{\lVert \hat{v}_t - v^\mathrm{r}_t \rVert^2}{\sigma\,\lVert v^\mathrm{c} \rVert^2}\right),
  \label{eq:reward}
\end{equation}
where $\sigma$ is the temperature.
The acceleration reward is the negative squared tracking error $-(\bar{\alpha}_t - \bar{g}_t)^2$, shifted such that a robot that stands still receives exactly zero and has to accelerate to get any reward.
However, at a constant walking speed, the acceleration is zero, which is why the hold reward additionally tracks the velocity reference with the velocity estimate.
The full reward $r_t = \Delta t\,(c_\mathrm{acc}\, r^\mathrm{acc}_t + c_\mathrm{hold}\, r^\mathrm{hold}_t + r^\mathrm{reg}_t) - c_\mathrm{fall}\,\mathds{1}^\mathrm{fall}_t$ adds a penalty for falling and the regularization terms $r^\mathrm{reg}_t$, which keep the yaw rate low, the trunk upright and the upper body close to its nominal pose and penalize fast changes in the actions.
The velocity estimate itself only enters the goal acceleration and the hold reward and is not part of the observation.

\subsection{Sample-Efficient Learning}
\label{sec:learning}

The position targets of the policy are tracked by Proportional-Derivative (PD) controllers.
To make the joints follow the targets closely, we use a high gain setting with a proportional gain of 60 and a derivative gain of 3.
With stiff and well-damped joints, the effect of every action becomes more predictable, which makes it easier to stabilize the robot early in training.
We use XQC~\citep{palenicek2025xqc} as our RL algorithm, which is an off-policy actor-critic algorithm that learns sample-efficiently through a well-conditioned critic optimization with batch normalization~\citep{bhatt2024crossq}, weight normalization and a distributional cross-entropy loss.

Naturally, a humanoid is left-right symmetric, and so is the locomotion task that we train on, so every collected transition has a mirrored counterpart that is equally valid~\citep{mittal2024symmetry}, i.e. taking a step to the left under a certain observation is equivalent to taking a step to the right under the mirrored observation.
We define a mirror operator $M$ that swaps the left and the right leg and negates all lateral observations and actions, and we replace every transition $(o_t, a_t, r_t, o_{t+1})$ sampled from the replay buffer by $(M o_t, M a_t, r_t, M o_{t+1})$ with a probability of 0.5.
Thereby, the policy and the critic learn about both sides of the gait from every collected transition with no additional cost.

\begin{figure}[t]
  \centering
  \includegraphics[width=\linewidth]{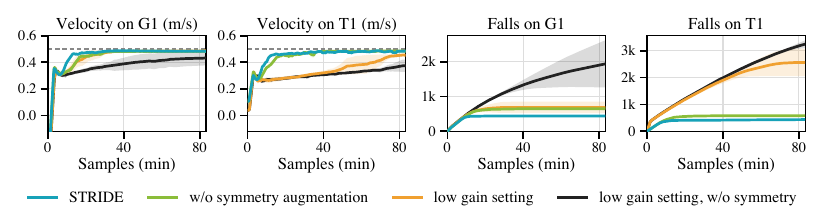}
  \caption{Forward velocity and number of falls during training for \method{} and its ablations, shown as mean and standard deviation over all runs. The dashed line shows the commanded velocity.}
  \label{fig:results}
\end{figure}

\begin{table}[t]
  \centering
  \caption{Runs that learn to walk within \budgetMin{} minutes, their average time to walk, and the average number of falls over all seeds.
  Every setting is trained on at least two seeds.}
  \label{tab:results}
  \footnotesize
  \setlength{\tabcolsep}{5pt}
  \begin{tabular}{lcccccc}
\toprule
 & \multicolumn{3}{c}{Unitree G1} & \multicolumn{3}{c}{Booster T1} \\
\cmidrule(lr){2-4}\cmidrule(lr){5-7}
Method & Walk & Time [min] & Falls & Walk & Time [min] & Falls \\
\midrule
\method{} & 2/2 & 13 & 443 & 2/2 & 13 & 437 \\
\quad w/o symmetry augmentation & 2/2 & 21 & 646 & 2/2 & 18 & 583 \\
\quad low gain setting & 4/4 & 22 & 689 & 2/2 & 66 & 2567 \\
\quad low gain setting, w/o symmetry & 4/7 & 49 & 1942 & 1/3 & 82 & 3250 \\
\midrule
low gain setting, w/o symmetry, UTD ratio 4 & 2/2 & 30 & 855 & 2/2 & 53 & 2108 \\
w/o symmetry, with CPG & 2/2 & 20 & 628 & 2/2 & 18 & 579 \\
\midrule
shared policy & 2/2 & 60 & 1760 & 0/2 & -- & 3792 \\
\bottomrule
\end{tabular}

\end{table}

\section{Experiments}
\label{sec:experiments}

We evaluate \method{} on the Unitree G1 and Booster T1 humanoids in the MuJoCo simulator~\citep{todorov2012mujoco} and implement in RL-X~\citep{bohlinger2024rlx}.
The policy controls the 12 leg joints at \controlHz{}\,Hz, and the robot is commanded to walk forward at 0.5\,m/s for 12\,s per episode.
To stay close to learning on a real robot, every run collects \budgetK{} transitions with a single robot, which corresponds to \budgetMin{} minutes worth of samples.
We report the time to walk and the falls during training, as they determine the cost of learning on a real robot.
A run learns to walk once its mean episode length reaches 500 of the 600 steps and its mean forward velocity reaches 0.4\,m/s in three consecutive windows of 1,000 steps.
The velocity of the simulator is only used for this evaluation.
We use $\alpha_\mathrm{max} = 0.75$\,m/s$^2$, $k_\mathrm{P} = 4$, $k_\mathrm{I} = 1$, $\tau = 0.25$\,s, $\sigma = 0.25$, $c_\mathrm{acc} = 10$, $c_\mathrm{hold} = 5$, $c_\mathrm{fall} = 2$ and a UTD ratio of 2.
For comparison, we also train without the symmetry augmentation and with a low gain setting, which uses PD gains of 45 and 1 for the G1 and of 25 and 0.6 for the T1, which are the default in RL-X.

Figure~\ref{fig:results} shows that \method{} reaches the commanded velocity on both humanoids within the first minutes of training.
Every run learns to walk after \minMinFull{} to \maxMinFull{} minutes worth of samples, which is earlier than any run of its ablations, and the final policies walk at \finalVelFull{}\,m/s and complete every episode.
Learning with the low gain setting and without symmetry augmentation is far less reliable, as only half of these runs learn to walk within the budget and the ones that do need \minDefault{} minutes on average (see Table~\ref{tab:results}).
Without the symmetry augmentation, the time to walk increases from \minFullGOne{} to \minNoSymGOne{} minutes on the G1 and to \minNoSymTOne{} minutes on the T1.
With the low gain setting, the G1 needs \minNoGainGOne{} minutes and the T1 even \minNoGainTOne{} minutes.
The number of falls stops increasing as soon as a policy walks, and across all runs that learn to walk, \fallsBeforeWalkMedianPct\% of the falls happen before that point.
Therefore, learning faster also means falling less, and \method{} reduces the average number of falls during training from over 2000 to \fallsFull{}.

Finally, a higher UTD ratio trades computational efficiency for sample efficiency.
Without the two components of \method{}, increasing it from 2 to 4 lets every run learn to walk and reduces the time to walk from \minDefaultGOne{} to \minUTDGOne{} minutes on the G1 and from \minDefaultTOne{} to \minUTDTOne{} minutes on the T1.
A CPG that generates alternating foot trajectories, on top of which the policy predicts residual joint targets~\citep{bellegarda2022cpgrl, bohlinger2025gaitineight}, does not change the time to walk with the high gain setting, so \method{} does not seem to benefit from a gait prior.
Furthermore, we train a single policy on both humanoids at once, which share the same observation and action space.
With the shared policy, every G1 run learns to walk, while the T1 does not yet learn to walk within the training budget.
In the current setup, we do not see positive transfer for training on both robots and sharing the collected transitions.
As the shared policy has to infer which robot it controls, embodiment-aware architectures~\citep{bohlinger2024onepolicy, bohlinger2025multiembodiment} are a natural next step to explore.

\section{Conclusion and Limitations}
\label{sec:conclusion}

We introduced \method{}, which learns humanoid locomotion from minutes of samples with a reward that only requires the onboard IMU.
High joint gains and symmetry augmentation turn a learning problem that only half of the runs solve within the budget into one that every run solves in a fraction of that time.
Our experiments are limited to simulation, where the acceleration of the IMU is free of noise and bias.
On a real robot, the IMU also measures gravity and its errors accumulate, so we expect that the hold reward will accumulate these errors as well.
In future work, we will bring \method{} to the physical G1 and T1, add omnidirectional commands and combine it with embodiment-aware network architectures to improve the transfer when training on both robots together.

\acknowledgments{
We acknowledge EuroHPC Joint Undertaking for awarding us access to LUMI at CSC, Finland, MeluXina at LuxProvide, Luxembourg, Vega at IZUM, Slovenia, and Arrhenius at NAISS, Sweden, through the Benchmark Access project EHPC-BEN-2026B08-075 for LUMI and the Development Access project EHPC-DEV-2026D08-135 for the other three systems.
LUMI is hosted by CSC (Finland) and the LUMI consortium.
The authors gratefully acknowledge the HPC RIVR consortium and EuroHPC JU for funding this research by providing computing resources of the HPC system Vega at the Institute of Information Science.
Some of the experiments were performed on the Luxembourg national supercomputer MeluXina, and the authors gratefully acknowledge the LuxProvide teams for their expert support.
Computational resources were provided by the National Academic Infrastructure for Supercomputing in Sweden (NAISS), funded by the Swedish Research Council.
The authors gratefully acknowledge the computing time granted by the KISSKI project.
The calculations for this research were conducted with computing resources under the project kisski-merl.
The authors gratefully acknowledge the computing time provided to them on the high-performance computer Lichtenberg II at the NHR Center NHR4CES at TU Darmstadt (project number p0028682).
This is funded by the Federal Ministry of Research, Technology and Space, and the state governments participating on the basis of the resolutions of the GWK for national high performance computing at universities.
Some of the experiments ran on the hessian.AI cluster 43, and we gratefully acknowledge support from the hessian.AI Service Center (funded by the Federal Ministry of Research, Technology and Space, BMFTR, grant no.~16IS22091) and the hessian.AI Innovation Lab (funded by the Hessian Ministry for Digital Strategy and Innovation, grant no.~S-DIW04/0013/003).
}

\bibliography{references}

\begin{thebibliography}{25}
\providecommand{\natexlab}[1]{#1}
\providecommand{\url}[1]{\texttt{#1}}
\expandafter\ifx\csname urlstyle\endcsname\relax
  \providecommand{\doi}[1]{doi: #1}\else
  \providecommand{\doi}{doi: \begingroup \urlstyle{rm}\Url}\fi

\bibitem[Hwangbo et~al.(2019)Hwangbo, Lee, Dosovitskiy, Bellicoso, Tsounis,
  Koltun, and Hutter]{hwangbo2019agile}
J.~Hwangbo, J.~Lee, A.~Dosovitskiy, D.~Bellicoso, V.~Tsounis, V.~Koltun, and
  M.~Hutter.
\newblock Learning agile and dynamic motor skills for legged robots.
\newblock \emph{Science Robotics}, 4\penalty0 (26), 2019.

\bibitem[Radosavovic et~al.(2024)Radosavovic, Xiao, Zhang, Darrell, Malik, and
  Sreenath]{radosavovic2024humanoid}
I.~Radosavovic, T.~Xiao, B.~Zhang, T.~Darrell, J.~Malik, and K.~Sreenath.
\newblock Real-world humanoid locomotion with reinforcement learning.
\newblock \emph{Science Robotics}, 2024.

\bibitem[Stasica et~al.(2025)Stasica, Bick, Bohlinger, Mohseni, Fritzsche,
  H{\"u}bler, Peters, and Seyfarth]{stasica2025bridge}
M.~Stasica, A.~Bick, N.~Bohlinger, O.~Mohseni, M.~J.~A. Fritzsche,
  C.~H{\"u}bler, J.~Peters, and A.~Seyfarth.
\newblock Bridge the gap: Enhancing quadruped locomotion with vertical ground
  perturbations.
\newblock In \emph{IEEE/RSJ International Conference on Intelligent Robots and
  Systems (IROS)}, 2025.
\newblock \doi{10.1109/IROS60139.2025.11247549}.

\bibitem[Rudin et~al.(2022)Rudin, Hoeller, Reist, and Hutter]{rudin2022minutes}
N.~Rudin, D.~Hoeller, P.~Reist, and M.~Hutter.
\newblock Learning to walk in minutes using massively parallel deep
  reinforcement learning.
\newblock In \emph{Conference on Robot Learning (CoRL)}, pages 91--100, 2022.

\bibitem[Bohlinger and Peters(2025)]{bohlinger2025epvf}
N.~Bohlinger and J.~Peters.
\newblock Massively scaling explicit policy-conditioned value functions.
\newblock \emph{arXiv preprint arXiv:2502.11949}, 2025.

\bibitem[Bohlinger et~al.(2024)Bohlinger, Czechmanowski, Krupka, Kicki, Walas,
  Peters, and Tateo]{bohlinger2024onepolicy}
N.~Bohlinger, G.~Czechmanowski, M.~P. Krupka, P.~Kicki, K.~Walas, J.~Peters,
  and D.~Tateo.
\newblock One policy to run them all: an end-to-end learning approach to
  multi-embodiment locomotion.
\newblock In \emph{Conference on Robot Learning (CoRL)}, volume 270 of
  \emph{Proceedings of Machine Learning Research}, pages 3356--3378, 2024.

\bibitem[Bohlinger and Peters(2025)]{bohlinger2025multiembodiment}
N.~Bohlinger and J.~Peters.
\newblock Multi-embodiment locomotion at scale with extreme embodiment
  randomization.
\newblock \emph{arXiv preprint arXiv:2509.02815}, 2025.

\bibitem[Bohlinger and Peters(2026)]{bohlinger2026active}
N.~Bohlinger and J.~Peters.
\newblock Active embodiment identification with reinforcement learning for
  legged robots.
\newblock \emph{arXiv preprint arXiv:2605.08020}, 2026.

\bibitem[Li et~al.(2026)Li, Ai, Bohlinger, Peters, Su, and
  Christensen]{li2026rapid}
D.~Li, B.~Ai, N.~Bohlinger, J.~Peters, H.~Su, and H.~I. Christensen.
\newblock Rapid embodiment adaptation for quadrupedal locomotion.
\newblock \emph{arXiv preprint arXiv:2608.01506}, 2026.

\bibitem[Smith et~al.(2023)Smith, Kostrikov, and Levine]{smith2023walk}
L.~Smith, I.~Kostrikov, and S.~Levine.
\newblock A walk in the park: Learning to walk in 20 minutes with model-free
  reinforcement learning.
\newblock In \emph{Robotics: Science and Systems (RSS)}, 2023.

\bibitem[Bohlinger et~al.(2025)Bohlinger, Kinzel, Palenicek, Antczak, and
  Peters]{bohlinger2025gaitineight}
N.~Bohlinger, J.~Kinzel, D.~Palenicek, {\L}.~Antczak, and J.~Peters.
\newblock Gait in eight: Efficient on-robot learning for omnidirectional
  quadruped locomotion.
\newblock In \emph{IEEE/RSJ International Conference on Intelligent Robots and
  Systems (IROS)}, pages 8790--8797, 2025.
\newblock \doi{10.1109/IROS60139.2025.11246447}.

\bibitem[Aditya et~al.(2025)Aditya, Huang, Bohlinger, Kicki, Walas, Peters,
  Luperto, and Tateo]{aditya2025robust}
D.~Aditya, J.~Huang, N.~Bohlinger, P.~Kicki, K.~Walas, J.~Peters, M.~Luperto,
  and D.~Tateo.
\newblock Robust localization, mapping, and navigation for quadruped robots.
\newblock In \emph{European Conference on Mobile Robots (ECMR)}, pages 1--8,
  2025.
\newblock \doi{10.1109/ECMR65884.2025.11163249}.

\bibitem[Haarnoja et~al.(2019)Haarnoja, Ha, Zhou, Tan, Tucker, and
  Levine]{haarnoja2019walk}
T.~Haarnoja, S.~Ha, A.~Zhou, J.~Tan, G.~Tucker, and S.~Levine.
\newblock Learning to walk via deep reinforcement learning.
\newblock In \emph{Robotics: Science and Systems (RSS)}, 2019.

\bibitem[Wu et~al.(2022)Wu, Escontrela, Hafner, Goldberg, and
  Abbeel]{wu2022daydreamer}
P.~Wu, A.~Escontrela, D.~Hafner, K.~Goldberg, and P.~Abbeel.
\newblock {DayDreamer}: World models for physical robot learning.
\newblock In \emph{Conference on Robot Learning (CoRL)}, 2022.

\bibitem[Li et~al.(2025)Li, Krause, and Hutter]{li2025robotic}
C.~Li, A.~Krause, and M.~Hutter.
\newblock Robotic world model: A neural network simulator for robust policy
  optimization in robotics.
\newblock \emph{arXiv preprint arXiv:2501.10100}, 2025.

\bibitem[Bhatt et~al.(2024)Bhatt, Palenicek, Belousov, Argus, Amiranashvili,
  Brox, and Peters]{bhatt2024crossq}
A.~Bhatt, D.~Palenicek, B.~Belousov, M.~Argus, A.~Amiranashvili, T.~Brox, and
  J.~Peters.
\newblock {CrossQ}: Batch normalization in deep reinforcement learning for
  greater sample efficiency and simplicity.
\newblock In \emph{International Conference on Learning Representations
  (ICLR)}, 2024.

\bibitem[Palenicek et~al.(2026)Palenicek, Vogt, Watson, Posner, and
  Peters]{palenicek2025xqc}
D.~Palenicek, F.~Vogt, J.~Watson, I.~Posner, and J.~Peters.
\newblock {XQC}: Well-conditioned optimization accelerates deep reinforcement
  learning.
\newblock In \emph{International Conference on Learning Representations
  (ICLR)}, 2026.

\bibitem[D'Oro et~al.(2023)D'Oro, Schwarzer, Nikishin, Bacon, Bellemare, and
  Courville]{doro2023replayratio}
P.~D'Oro, M.~Schwarzer, E.~Nikishin, P.-L. Bacon, M.~G. Bellemare, and
  A.~Courville.
\newblock Sample-efficient reinforcement learning by breaking the replay ratio
  barrier.
\newblock In \emph{International Conference on Learning Representations
  (ICLR)}, 2023.

\bibitem[Smith et~al.(2024)Smith, Cao, and Levine]{smith2024grow}
L.~Smith, Y.~Cao, and S.~Levine.
\newblock Grow your limits: Continuous improvement with real-world {RL} for
  robotic locomotion.
\newblock In \emph{IEEE International Conference on Robotics and Automation
  (ICRA)}, pages 10829--10836, 2024.

\bibitem[Mittal et~al.(2024)Mittal, Rudin, Klemm, Allshire, and
  Hutter]{mittal2024symmetry}
M.~Mittal, N.~Rudin, V.~Klemm, A.~Allshire, and M.~Hutter.
\newblock Symmetry considerations for learning task symmetric robot policies.
\newblock In \emph{IEEE International Conference on Robotics and Automation
  (ICRA)}, pages 7433--7439, 2024.

\bibitem[Bellegarda and Ijspeert(2022)]{bellegarda2022cpgrl}
G.~Bellegarda and A.~Ijspeert.
\newblock {CPG-RL}: Learning central pattern generators for quadruped
  locomotion.
\newblock \emph{IEEE Robotics and Automation Letters}, 2022.

\bibitem[Ai et~al.(2025)Ai, Dai, Bohlinger, Li, Mu, Wu, Fay, Christensen,
  Peters, and Su]{ai2025embodiment}
B.~Ai, L.~Dai, N.~Bohlinger, D.~Li, T.~Mu, Z.~Wu, K.~Fay, H.~I. Christensen,
  J.~Peters, and H.~Su.
\newblock Towards embodiment scaling laws in robot locomotion.
\newblock In \emph{Conference on Robot Learning (CoRL)}, volume 305 of
  \emph{Proceedings of Machine Learning Research}, pages 3483--3515, 2025.

\bibitem[Bohlinger and Peters(2026)]{bohlinger2026shape}
N.~Bohlinger and J.~Peters.
\newblock Shape your body: Value gradients for multi-embodiment robot design.
\newblock In \emph{Conference on Robot Learning (CoRL)}, 2026.

\bibitem[Todorov et~al.(2012)Todorov, Erez, and Tassa]{todorov2012mujoco}
E.~Todorov, T.~Erez, and Y.~Tassa.
\newblock {MuJoCo}: A physics engine for model-based control.
\newblock In \emph{IEEE/RSJ International Conference on Intelligent Robots and
  Systems (IROS)}, 2012.

\bibitem[Bohlinger and Dorer(2024)]{bohlinger2024rlx}
N.~Bohlinger and K.~Dorer.
\newblock {RL-X}: A deep reinforcement learning library (not only) for
  {RoboCup}.
\newblock In \emph{RoboCup 2023: Robot World Cup XXVI}, volume 14140 of
  \emph{Lecture Notes in Computer Science}, pages 228--239. Springer, 2024.

\end{thebibliography}

\end{document}